\documentclass[10pt,twocolumn,letterpaper]{article}

\usepackage{cvpr} 
\usepackage{times}
\usepackage{epsfig}
\usepackage{graphicx}
\usepackage{amsmath}
\usepackage{amssymb}
\usepackage{booktabs}
\usepackage{algorithm}
\usepackage{algorithmic}
\usepackage{colortbl}
\usepackage{multirow}
\usepackage{xcolor}
\usepackage{hyperref}
\definecolor{lightblue}{rgb}{0.8,0.88,1}
\definecolor{lightred}{rgb}{1,0.8,0.8}
\definecolor{lightgray}{gray}{0.95}

\begin{document}

\title{DArFace: Deformation Aware Robustness for Low Quality Face Recognition}

\author{
Sadaf Gulshad \quad Abdullah Aldahlawi\\
Advanced AI and Information Technology, Thakaa\\
Riyadh, Saudi Arabia\\
{\tt\small sadafgulshad@gmail.com, aldahlawi@thakaait.net}
}
\maketitle
\thispagestyle{empty}

\begin{abstract}
  Facial recognition systems have achieved remarkable success by leveraging deep neural networks, advanced loss functions, and large-scale datasets. However, their performance often deteriorates in real-world scenarios involving low-quality facial images. Such degradations, common in surveillance footage or standoff imaging include low resolution, motion blur, and various distortions, resulting in a substantial domain gap from the high-quality data typically used during training. While existing approaches attempt to address robustness by modifying network architectures or modeling global spatial transformations, they frequently overlook local, non-rigid deformations that are inherently present in real-world settings. In this work, we introduce \textbf{DArFace}, a \textbf{D}eformation-\textbf{A}ware \textbf{r}obust \textbf{Face} recognition framework that enhances robustness to such degradations without requiring paired high- and low-quality training samples. Our method adversarially integrates both global transformations (e.g., rotation, translation) and local elastic deformations during training to simulate realistic low-quality conditions. Moreover, we introduce a contrastive objective to enforce identity consistency across different deformed views. Extensive evaluations on low-quality benchmarks including TinyFace, IJB-B, and IJB-C demonstrate that DArFace surpasses state-of-the-art methods, with significant gains attributed to the inclusion of local deformation modeling.
  
\end{abstract}

\section{Introduction}
\label{sec:intro}
\begin{figure}[t]
\centering
   \includegraphics[width=\linewidth]{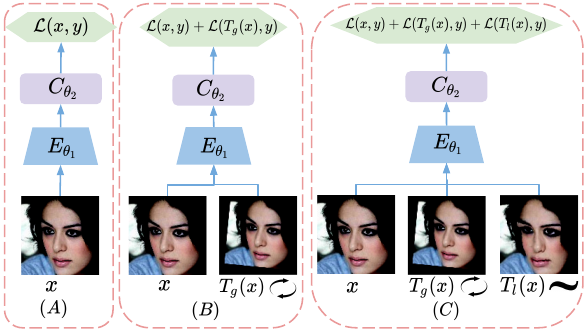}
   \caption{\textbf{Comparison of our method with the current methods:} (A) clean-only training uses only original images $x$, (B) \cite{saadabadi2024ARoFace} uses clean + globally transformed $T_g(x)$ images for training, (C) Our method (\textbf{DArFace}) uses \textit{clean}, \textit{globally $T_g(x)$ (e.g., rotated)}  and \textit{locally $T_l(x)$ (e.g., elastic deformed eye and lip region)} transformed images for improved robustness.}
\label{fig:teaser}
\end{figure}
 In the last few years, facial recognition methods have achieved remarkable performance, their success is largely attributed to the advent of deep neural networks ~\cite{he2016deep, krizhevsky2012imagenet,huang2017densely}, the design of sophisticated loss functions ~\cite{liu2017sphereface, wang2018cosface, deng2019ArcFace, kim2022adaface,hoffer2015deep}, and training on large-scale datasets ~\cite{guo2016ms,zhu2021webface260m, huang2008labeled, kemelmacher2016megaface}. However, it has turned out that not all images encountered in the real world scenarios are of good quality. For example, images captured from a standoff distance e.g. surveillance cameras often suffer from low resolution, motion blur, or camera artifacts, resulting in low quality images. The state of the art methods tend to perform worst on such low quality inputs compared to high quality ones \cite{saadabadi2024ARoFace, liu2022controllable,cheng2018low,zangeneh2020low}. This performance drop is mainly attributed to the domain gap between the high quality well curated datasets on which usually models are trained and low quality test sets ~\cite{saadabadi2024ARoFace, zangeneh2020low, wang2019improved, singh2019dual, shi2021boosting, liu2022controllable, ge2020efficient}. A straightforward solution is to collect a large-scale low-quality dataset and train models directly on it. However, this approach faces several challenges, including noisy labels, privacy concerns and the high costs for dataset acquisition.

The facial recognition pipeline can be broadly divided into three stages: face detection, alignment and recognition. In the context of low-quality inputs, existing approaches can be grouped based on which part of the pipeline they target. Some works address the effects of degradation on detection and alignment ~\cite{saadabadi2024ARoFace, singh2019dual}, while others focus on improving robustness during recognition ~\cite{liu2022controllable, shin2022teaching, wang2019improved, yang2021gan, singh2019dual}.  Recognition focused models either attempt to map both low and high quality inputs into the shared space or aim to estimate high quality representations from their low quality counter parts. However, the  former approach lacks practicality due to the scarcity of low quality training data, while the latter faces the challenge of ambiguity, several plausible high quality counterparts exist for a single low quality input. 

Recently \cite{singh2019dual} utilized instantiation parameters such as pose, lighting, and deformation in capsule networks \cite{hinton2011transforming}. They argued that these parameters can make the network more robust to the variations in resolution of the image, thereby improving the performance on low quality inputs. In an orthogonal direction, \cite{saadabadi2024ARoFace} investigated the impact of alignment error on the recognition performance. Rather than modifying the architecture they employed global spatial transformations to simulate the misalignment and incorporated these adversarially perturbed samples into the training, thereby enhancing robustness to the alignment noise. 

While \cite{saadabadi2024ARoFace} focuses solely on the impact of global spatial transformations such as translation, rotation and scaling, our work extends this line of work by incorporating local non rigid differentiable deformations into the training process, see Figure. \ref{fig:teaser}. These local elastic deformations resemble real-world degradations caused by lens distortions, compression artifacts, motion blur and subtle facial muscle movements, factors that are not captured by the global transformations alone. We propose a deformation-aware training strategy that combines both local and global transformations in an adversarial training setup. This enables the network to learn the features that are not only robust to the alignment errors but also to more localized and content-preserving distortions. Furthermore, we employ contrastive objective to ensure that the embeddings stay stable across different deformations and focuses on the identity rather than deformation type, thus improving performance on low quality inputs at the test time.
\begin{enumerate}
    \item We propose \textbf{DArFace}, a novel deformation-aware training framework for low-quality face recognition that simulates realistic degradations without requiring paired high- and low-quality data.
    \item Our method adversarially integrates both global transformations (e.g., rotation, translation) and local elastic deformations during training to improve robustness against real-world variations.
    \item We introduce a contrastive identity loss that enforces feature alignment between different deformed views, promoting deformation-invariant face embeddings.
    \item Without relying on architectural modifications, DArFace achieves consistent and state-of-the-art improvements across multiple low-quality benchmarks, including TinyFace, IJB-B, and IJB-C.
\end{enumerate}

\begin{figure*}[htbp]
    \centering
    \includegraphics[width=\linewidth]{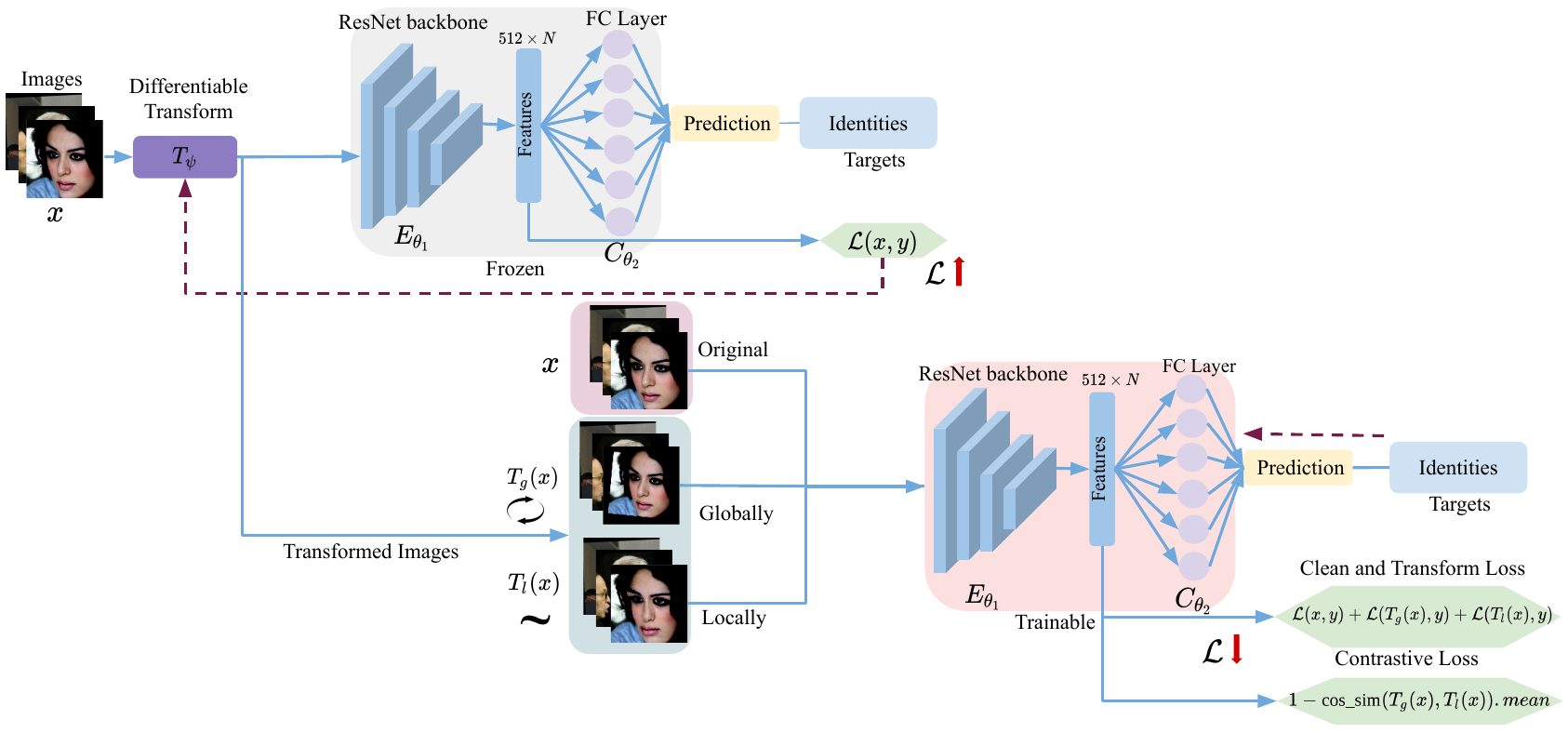}
    \caption{
\textbf{Overview of DArFace training framework.} 
Step1 (Top Row- Adversarial Search): A clean image $x$ is transformed $T_{\psi}$ and passed through a frozen network to search for global and local  parameters such that the loss $\mathcal{L}$ maximally increases. Step2 (Bottom Row-Forward Pass): Clean, globally (e.g. rotated) and locally (warped eye and lip) transformed $(x, T_g(x),T_l(x))$ inputs are fed to the model for training. Step3 (Bottom Row-Backward Pass): We minimize the total loss $\mathcal{L}_{clean}+\mathcal{L}_{trans}+\lambda_{cont}\mathcal{L}_{cont}$, i.e. angular loss on clean and transformed views and a contrastive loss that aligns global and local feature pairs.
}
    \label{fig:architecture}
\end{figure*}

\section{Related Work}
\label{sec:relatedwork}
\subsection{Low-Quality Face Recognition}
With the recent success of deep neural networks, facial recognition models have achieved remarkable performance on datasets where facial attributes are clearly perceptible, such as LFW, CFP-FP, CPLFW, AgeDB and CALFW ~\cite{huang2008labeled, sengupta2016frontal, zheng2018cross, moschoglou2017agedb, zheng2017cross}. However, when these models are directly applied to the low quality images such as IJB-B, IJB-C, IJB-S and TinyFace ~\cite{whitelam2017iarpa, maze2018iarpa, kalka2018ijb, cheng2018low} as those affected by low resolution, motion blur, or sensor noise a significant performance drop is observed \cite{saadabadi2024ARoFace, kim2022adaface, cheng2018low}. This degradation is primarily attributed to the domain gap between the good quality training data and low quality test data \cite{cheng2018low, zou2011very}. Furthermore, training directly on low quality inputs is not optimal, as such inputs often lack sufficient visual information for representation learning. 

A straight forward approach to mitigate this issue is to collect low quality dataset for training. However, this direction faces several practical challenges, including privacy concerns, noisy labels and high costs for data collection and annotation ~\cite{cheng2018low, liu2022controllable}. An  alternative solution is to generate or synthesize high quality counterparts for the low quality inputs ~\cite{liu2022controllable, shin2022teaching, yang2021gan, Li_2020_CVPR, Wang_2022_CVPR}. However, this technique suffers from ill-posedness i.e. multiple high quality outputs corresponds to a single low quality input ~\cite{Huang_2017_ICCV, shin2022teaching, 8723565, 10.1007/978-3-319-10593-2_25}. 

Other approaches attempt to project high quality and low quality images into a shared space~\cite{wang2019improved, shin2022teaching, ge2020efficient, ZANGENEH2020112854}. While promising, this direction still relies on access to low quality inputs during training, which may not always be possible in the real world applications. In another direction, quality aware fusion approaches have been proposed \cite{Chang_2020_CVPR,Li_2021_CVPR,5204299}, often using probabilistic approaches to estimate the uncertainty in the recognition process. These approaches selectively learn features from the informative samples and those with very few identity cues are discarded. A notable example is \cite{kim2022adaface}, which modifies the loss function and uses feature norm as a proxy feature quality to guide training. However, these approaches either resort to learning mean and variance separately or rely on the availability of the low quality data.

In parallel, another group of work focuses on architectural and training based modifications to enhance the robustness against common distortions in low quality images ~\cite{singh2019dual,zangeneh2020low, ge2020efficient}. Capsule networks~\cite{zangeneh2020low},  have been used to encode pose, lighting, and deformations as instantiation parameters, making the network resilient to resolution and structural changes. On the other hand, transformation aware training strategies like  ~\cite{saadabadi2024ARoFace} introduces global transformations like scaling and rotation to mitigate misalignment and enhance the robustness without architectural changes.  

While our approach aims to combine differentiable global spatial transformations with the local, non rigid deformations within the training process.
This combination allows for learning representations that are robust to both alignment errors and subtle content-preserving distortions common in low quality facial images. We further incorporate contrastive loss term with a standard angular margin loss to encourage the network to learn identity consistent features across global and local transformations.

\subsection{Adversarial Augmentations}
Despite the remarkable performance of deep neural networks, \cite{szegedy2013intriguing} demonstrated that the small, carefully crafted perturbations known as adversarial perturbations can lead to incorrect predictions. Since then adversarial perturbations and adversarial training (to enhance the robustness) have been extensively explored in the computer vision \cite{goodfellow2014explaining, szegedy2013intriguing, kurakin2018adversarial,papernot2016limitations,moosavi2016deepfool}. Adversarial perturbations can be broadly categorized into three types: 1) \textit{intensity based perturbations,} involves subtle modifications to pixel intensities \cite{szegedy2013intriguing, goodfellow2014explaining, papernot2016limitations}, 2) \textit{geometric perturbations,} includes differentiable transformations e.g. rotation, scaling, and translation \cite{xiao2018spatially, saadabadi2024ARoFace} and 3) \textit{natural perturbations,}  mimics real world modifications e.g. adding makeup to faces, wearing glasses etc \cite{sharif2016accessorize,komkov2021advhat,deb2020advfaces,liu2022controllable}. 

In the realm of facial recognition \cite{deb2020advfaces} utilized GANs to generate minimal perturbations in the salient face regions. \cite{sharif2016accessorize} constrained adversarial regions to eye glasses, while \cite{sharif2019general} synthesized adversarial glasses using GANs. \cite{yang2021attacks} employed attention mechanism and variational autoencoders to misclassify individuals as target persons. \cite{liu2022controllable} generated facial images in a controllable manner and incorporated them in the adversarial training to enhance the network robustness. However, these methods often require predefined target data and are tailored for specific applications. Closely related to our work, ~\cite{saadabadi2024ARoFace} used global spatial transformations to mitigate the alignment errors in the facial recognition system. We focus on integrating local transformations along with the global ones to enhance the low quality recognition performance.

\subsection{Contrastive Learning for Facial Recognition}
Contrastive learning in facial recognition aims to improve intra-class compactness and enhance inter-class separability \cite{schroff2015facenet}. Since the introduction of triplet loss in \cite{schroff2015facenet}, contrastive learning has played a pivotal role in advancing facial recognition ~\cite{deng2019ArcFace, wang2018cosface,liu2017sphereface,kim2022adaface}. \cite{deng2019ArcFace} introduced an angular margin to the softmax loss to better separate identities in the angular embedding space. Similarly ~\cite{wang2018cosface} imposed a cosine margin to further improve the decision boundary in the angular space. In our work, we combine global and local deformations within an adversarial training framework. To enforce identity-consistent features across different perturbations, we introduce an additional contrastive loss alongside the angular margin loss. This encourages the model to focus on identity rather than transformation-specific features, thereby improving robustness to low-quality variations.

\section{Proposed Method}
\subsection{Notations}

Let $\mathcal{D}=\{x_n,y_n\}_{n=1}^N$ be the training dataset, consisting of $N$ high-resolution face images. Each image $x \in \mathbb{R}^{H \times W \times 3}$ corresponds to an identity label $y \in \{1, \dots, I\}$, where $I$ is the total number of identities. The face recognition model is $F_\theta = C\cdot E$, where $E_{\theta_1}(\cdot):\mathbb{R}^q \rightarrow \mathbb{R}^d$ is the feature extractor and $C_{\theta_2}(\cdot):\mathbb{R}^d \rightarrow \mathbb{R}^c$ is the classifier, usually a hyperspherical classifier \cite{deng2019ArcFace, wang2018cosface}.
Let $T_g(\cdot)$ denote a global transformation function (e.g., rotation, scaling, translation), and $T_l(\cdot)$ be a local deformation function (e.g., elastic distortion).

\subsection{Preliminaries}
We propose to leverage both local and global transformations for enhancing the robustness and ultimately improving the performance on low quality image recognition, as shown in Figure \ref{fig:architecture}. First, a clean image $x$ is transformed $T_{\psi}$ and passed through a frozen network to search for global and local parameters such that the loss $\mathcal{L}$ maximally increases. This yields adversarial global and local views, $T_g(x)$ and $T_l(x)$, respectively. Next, the clean, globally, and locally transformed images $(x, T_g(x), T_l(x))$ are jointly used for training. The model is optimized using a composite loss: $\mathcal{L}_{clean}+\mathcal{L}_{trans}+\lambda_{cont}\mathcal{L}_{cont}$, , where $\mathcal{L}_{clean}$ and $\mathcal{L}_{trans}$ are angular losses on the clean and transformed views, and $\mathcal{L}_{cont}$ is a contrastive loss that encourages alignment between global and local feature pairs. Our method integrates adversarial data augmentation by introducing differentiable global and local deformations into the training process. In the following sections we first describe the adversarial training framework and then outline the types of transformations used. Finally we present the details of our proposed method.

\noindent\textbf{Adversarial Data Augmentation} is applied to enhance the robustness of neural networks by training on perturbed inputs alongside the clean ones \cite{goodfellow2014explaining, madry2017towards}. This process usually follows two step min-max formulation: \textit{1) inner maximization}, i.e. generating adversarial examples to maximize the loss:

\begin{equation}  
\psi^* = \arg\max_{\psi} \mathcal{L}(F_{\theta}, T_{\psi}(x), y)
\end{equation}
\noindent where \( T_{\psi} \) is a transformation conditioned on the parameter \( \psi \).

\textit{2) outer minimization}, i.e. model is trained on both adversarial and clean inputs to minimize the loss:
\begin{equation}
\theta^* = \arg\min_{\theta} \mathbb{E}_{(x, y) \sim \mathcal{D}} \left[ \mathcal{L}(F_\theta(x), y) + \lambda \, \mathcal{L}(F_\theta(T_{\psi^*}(x)), y) \right]
\end{equation}
\noindent where \( \lambda \) controls the balance between clean and adversarial loss terms. The parameter \( \psi \) is projected onto an \( \ell_p \)-norm ball \( \mathcal{C} \) to preserve perceptual similarity.

 \begin{algorithm}[t]
\caption{DArFace Training}
\label{alg:adv_train}
\begin{algorithmic}[1]
\STATE \textbf{Input:} Dataset $\mathcal{D}$, model $F_\theta$, transformation parameters $\psi_0$, learning rate $\eta$, adversarial step size $\eta_\psi$, number of ascent steps $K$, projection bounds $\mathcal{C}$, trade-off coefficient $\lambda_{\text{cont}}$
\FOR{mini-batch $(x,y) \sim \mathcal{D}$}
    \STATE \textbf{Adversarial search:} $\psi \leftarrow \psi_0$
    \FOR{$k=1$ to $K$}
        \STATE $\psi \leftarrow \text{Proj}_{\mathcal{C}}\!\left(\psi + \eta_\psi \operatorname{sgn}\big(\nabla_\psi \mathcal{L}(F_\theta(T_\psi(x)), y)\big)\right)$
    \ENDFOR
    \STATE \textbf{Losses:}
    \STATE $\mathcal{L}_{\text{clean}} \leftarrow \mathcal{L}(F_\theta(x), y)$,\quad
           $\mathcal{L}_{\text{trans}} \leftarrow \mathcal{L}(F_\theta(T_\psi(x)), y)$
    \STATE Form global/local views $X_{{g}}, X_{{l}}$; get features $F_{{g}}=F_\theta(X_{{g}})$, $F_{{l}}=F_\theta(X_{{l}})$
    \STATE $\mathcal{L}_{\text{cont}} \leftarrow 1 - \text{cos\_sim}(F_{{g}}, F_{{l}}).\text{mean()}$
    \STATE \textbf{Update:} $\theta \leftarrow \theta - \eta \nabla_\theta\!\big(\mathcal{L}_{\text{clean}} + \mathcal{L}_{\text{trans}} + \lambda_{\text{cont}} \mathcal{L}_{\text{cont}}\big)$
\ENDFOR
\end{algorithmic}
\end{algorithm}

\noindent\textbf{Differentiable Global and Local Transformations} are applied to enhance robustness against pose variations and fine-grained spatial distortions in low-quality inputs. Let the input image be denoted as $x\in\mathbb{R}^q$, where $q=h\times w\times 3$. Let $(i,j)$ be the row and column indices of the image and $(u,v)\in\mathbb{R}^2$ are the horizontal and vertical axis. pixel indices are mapped to centered coordinate by $P_u(j)=j-\frac{w-1}{2}$, $P_v(j)=\frac{h-1}{2}-i$. Both the global $T_g$ and local $T_l$ transforms are invertible and parameterized by $\psi=(\varphi,\Delta u,\Delta v, \lambda, \alpha,\sigma)$, where $\varphi \in [0,2\pi]$ is rotation angle, $\Delta u \in \mathbb{R}$, $\Delta v \in \mathbb{R}$ denote horizontal and vertical shifts respectively, $\lambda \in \mathbb{R}$ denotes scaling factor, $\alpha \in \mathbb{R}_+$ is the elasticity coefficient and $\sigma\in \mathbb{R}_+$ is the smoothing parameter. Each pixel coordinate $(P_u(j),P_v(i))$ is transformed by:
\begin{equation}
    (u',v') =T^{-1}_{\psi}(P_u(j),P_v(i))
\end{equation}

The corresponding pixel value is then obtained using bilinear interpolation:
\begin{equation}
x'_{i,j} = I_x(u', v')
\end{equation}

where \( I_x(\cdot) \) is the bilinear interpolation function defined as:
\begin{align}
I_x(u, v) = \sum_{i=0}^{h-1} \sum_{j=0}^{w-1} x_{i,j} \cdot 
&\max(0, 1 - |v - P_v(i)|) \notag \\
&\cdot \max(0, 1 - |u - P_u(j)|)
\label{eq:bilinear_interp}
\end{align}

This gives the transformed pixel value at each location: 
\begin{align}
    x'_{i,j}=I_x(T^{-1}_{\psi}(P_u(j),P_v(i)))
\end{align}

\noindent\textbf{Local Elastic Deformations} are used to model spatial distortions through a two stage process \cite{1227801}. At first local affine transformations $A_\psi$ are applied to the input. To estimate the transformation parameters, three control points in the input grid are randomly perturbed using a uniform distribution:
\begin{equation}
(u_t, v_t) = (u + U(-\alpha, \alpha),\; v + U(-\alpha, \alpha))
\end{equation}
Affine transform is applied to get the target coordinates
\begin{equation}
\begin{bmatrix}
u_t \\
v_t \\
1
\end{bmatrix}
= 
A_{\psi}
\begin{bmatrix}
u \\
v \\
1
\end{bmatrix}
\end{equation}

Secondly, Gaussian warping is applied to each coordinate \( (u, v) \) by adding a smooth, spatially varying displacement defined by a Gaussian kernel is added:

\begin{equation}
  u' = u + \alpha \cdot \frac{1}{\sqrt{2\pi}\sigma} \exp\left(-\frac{u^2}{2\sigma^2}\right)  
\end{equation}
\begin{equation}
   v' = v + \alpha \cdot \frac{1}{\sqrt{2\pi}\sigma} \exp\left(-\frac{v^2}{2\sigma^2}\right) 
\end{equation}

This introduces a deformed grid \( (u', v') \), which captures elastic displacements in both spatial dimensions, while preserving differentiability.

\subsection{Deformation-Aware Augmentation}
Our aim is to robustify facial recognition network by enriching the training with both global transforms as well as local transforms.   The transformations are fully differentiable making it possible to integrate them in the adversarial training framework.
\begin{table*}
  \centering
\begin{tabular}{c c c c c c}
\toprule
\rowcolor{lightgray}
 \textbf{Method} &  \textbf{LFW} & \textbf{CFP-FP} & \textbf{AgeDB30} & \textbf{CALFW} & \textbf{CPLFW}\\ 
\midrule
Softmax & 99.70 & 98.20 & 97.72 & 95.65 & 92.02\\
ArcFace \cite{deng2019ArcFace}& 99.76 & 98.54 & \textbf{98.28} & 96.10 & 93.16\\
MagFace \cite{meng2021magface} & \textbf{99.83} & 98.46 & 98.17 & 96.15 & 92.87\\
ARoFace \cite{saadabadi2024ARoFace}&  99.78 & 98.51 & 97.80 & 95.80 & 93.43\\
\rowcolor{lightred}
\textbf{DArFace (Ours)} & 99.80 & \textbf{98.62} & 98.06 & \textbf{96.20} & \textbf{93.64} \\

\bottomrule
\end{tabular}
\vspace{4pt}
\caption{ \textbf{Performance comparison on high quality inputs using a ResNet-100 backbone and MS1MV2 training dataset.} DArFace achieves comparable or superior performance on standard high-quality datasets, demonstrating its robustness even in the absence of distortions or low-quality artifacts at test time.}
\label{tab:highquality}
\end{table*}

\begin{table*}[h]
  \centering
  \small
  \setlength{\tabcolsep}{6pt}
  \resizebox{0.95\textwidth}{!}{
  \begin{tabular}{lccccccccccc}
    \toprule
    \rowcolor{lightgray}
    \textbf{Method} & \textbf{Train Set}& \multicolumn{5}{c}{\textbf{IJB-B}} & \multicolumn{5}{c}{\textbf{IJB-C}}\\ 
    \rowcolor{lightgray}
     & & \multicolumn{3}{c}{\textbf{TAR@FAR}} & \multicolumn{2}{c}{\textbf{Identification}} & \multicolumn{3}{c}{\textbf{TAR@FAR}} & \multicolumn{2}{c}{\textbf{Identification}} \\
    \rowcolor{lightgray}
    & &\textbf{1e-6}& \textbf{1e-5}& \textbf{1e-4}& \textbf{Rank 1}& \textbf{Rank 5} &\textbf{1e-6}& \textbf{1e-5}& \textbf{1e-4}&\textbf{Rank 1}& \textbf{Rank 5} \\
    
    \midrule
    MagFace \cite{meng2021magface} & MS1MV2 & 40.57 & 89.84 & 94.35 &-- & -- & 89.10 & 93.70 &95.81  & -- & --\\
    MagFace+IIC \cite{huang2024enhanced} & MS1MV2 & -- & -- & -- &-- & --  & 89.38 & 93.95 &95.89 & -- & --\\
    ArcFace \cite{deng2019ArcFace} & MS1MV2 & 43.15 & 88.42  & 94.53 & 94.71 & 96.66 & 85.12 & 93.28 &95.82 & 95.99 & 97.16\\
    ArcFace + CFSM \cite{liu2022controllable} & MS1MV2 & 47.27 & \textbf{90.52} & \textbf{95.21} & \textbf{95.00} &96.67 & \textbf{90.83} & \textbf{94.72}&\textbf{96.60} & \textbf{96.19} & 97.30\\
    ARoFace \cite{saadabadi2024ARoFace} & MS1MV2 & 46.24 & 85.28 & 94.09 & 94.44& 96.49 & 79.59 & 91.05 & 95.46 & 95.58 & 97.01 \\
    \rowcolor{lightred}
    \textbf{DArFace (Ours)} & MS1MV2 & \textbf{48.08} & {87.11} & {94.40} & {94.82} & \textbf{96.82} & {84.83} & {93.01} &  {95.87} & {96.00} & \textbf{97.32} \\
    
    \midrule
    ArcFace+VPL \cite{deng2021variational} & MS1MV3 & -- & -- & \textbf{95.56}& -- & -- & -- & --&\textbf{96.67} & -- & -- \\
    ArcFace+SC \cite{deng2020sub} & MS1MV3 & -- & -- &95.25& -- & -- & -- & --&96.61 & -- & -- \\
    ArcFace \cite{deng2019ArcFace} & MS1MV3 & 37.15 & 89.81 & 94.77 &95.12 & 96.92 & 85.20 & 93.26& 95.83 & 96.01 & 97.17 \\
    ARoFace \cite{saadabadi2024ARoFace} & MS1MV3 & 39.36 & 90.17 & 95.09& 95.21 & 97.13 & \textbf{87.00} & 94.12 & 96.34 & 96.57 &97.72 \\
    \rowcolor{lightred}
    \textbf{DArFace (Ours)} & MS1MV3 & \textbf{44.04} & \textbf{90.32} & {95.11}& \textbf{95.33} & \textbf{97.19} & {86.71} & \textbf{94.14} &{96.40}& {96.42} & {97.64} \\
    \midrule
    ArcFace \cite{deng2019ArcFace} & WebFace4M & 42.19 & 87.60 & 94.43 &95.12 & 96.83 & 69.03 & 89.40& 95.70 & 96.03 & 97.50 \\
    ARoFace \cite{saadabadi2024ARoFace} & WebFace4M & \textbf{44.88} & 91.82 & 95.43& 95.78 & 97.41 & 87.90 & 94.58 & 96.85 & 96.99 &98.06 \\
    AdaFace \cite{kim2022adaface} & WebFace4M & 44.48 & \textbf{92.26} & \textbf{96.03} & \textbf{96.26} & \textbf{97.68} &\textbf{ 90.43} & \textbf{95.34} & \textbf{97.39} & \textbf{97.52} &\textbf{98.35} \\
    \rowcolor{lightred}
    \textbf{DArFace (Ours)} & WebFace4M & 44.33 & 92.06 & 95.45 & 95.79 & 97.43 & 88.93 & 94.76 &  96.82 & 97.10 & 98.05 \\
    \midrule
    ArcFace \cite{deng2019ArcFace} & WebFace12M & 41.96 & 88.18 & 95.24 &95.45 & 97.15 & 74.25 & 91.02& 96.46 & 96.45 & 97.70 \\
    ARoFace \cite{saadabadi2024ARoFace} & WebFace12M & {44.61} & 93.05 & 96.14& 96.09 & 97.48 & \textbf{89.51} & 95.62 & 97.35 & 97.25 &98.15 \\
    AdaFace \cite{kim2022adaface} & WebFace12M & 47.49 & \textbf{93.13} & \textbf{96.30} & \textbf{96.28} & \textbf{97.72} &{ 89.47} & \textbf{95.94} & \textbf{97.54} & \textbf{97.56} &\textbf{98.38} \\
    \rowcolor{lightred}
    \textbf{DArFace (Ours)} & WebFace12M & \textbf{48.37} & 92.73 & 96.23 & 96.25 & 97.59 & 86.34 & 95.23 &  97.41 & 97.32 & 98.19 \\
    \bottomrule
  \end{tabular}}
  \vspace{4pt}
  \caption{\textbf{Performance comparison on mixed-quality inputs (IJB-B and IJB-C) using a ResNet-100 backbone and MS1MV2, MS1MV3, WebFace4M and WebFace12M training dataset.} DArFace consistently outperforms prior baselines, particularly in low-FAR and identification metrics, highlighting its robustness against quality degradations. Note that, ArcFace+CFSM, a GAN-based approach, relies on synthetic data generation and careful tuning, making its training process more complex. }
  \label{tab:medquality}
\end{table*}
\noindent\textbf{Training.} The training procedure follows two steps, as outlined in Algorithm \ref{alg:adv_train}.  In the first step, we optimize the transformation parameters $\psi$  such that the loss of the facial recognition network $\mathcal{L}$ increases. These transformations include global transforms like rotation $\varphi$ which rotates the input, and local non rigid transforms like elastic smoothing factor $\sigma$. After finding the transformation parameters $\psi^*$ that increases the loss the most within a bounded range $\mathcal{C}$, the second step updates the network parameters $\theta$ to minimize the classification loss over both clean images and the adversarially transformed images. The training can be formulated as: 
\begin{equation}
    \arg\min_{\theta} \frac{1}{\mathcal{D}}\sum_{(x,y)\in \mathcal{D}}[\mathcal{L}(F_\theta;x,y)+\arg\max_\psi\mathcal{L}((F_\theta;T_{\psi}(x),y))]
\end{equation}
where $\mathcal{L}$ in our work is ArcFace loss and $T_{\psi}$ are both global and local transforms.

Unlike the traditional adversarial example generation methods that modify the input by changing the pixel values directly, our method perturbs the global and local transformation parameters. For instance, modifying $\varphi$ rotates the input image, while updating $\sigma$ introduces local smooth elastic distortions. These perturbations are content-preserving and simulate real-world degradations such as misalignment, compression artifacts, and motion blur.

In order to generate realistic and diverse deformations the parameter values are randomly sampled from Gaussian distributions. These sampled parameters are then applied to the input images and the network is encouraged to learn features that are invariant to a broad range of real-world deformations.

\noindent\textbf{Loss.} While adversarial deformations make the model robust to various geometric distortions, they can also lead to instability in the feature space. Different transformations of the same identity might result in divergent embeddings, leading to potential identity mismatches. To address this, we introduce a contrastive loss that enforces the consistency between images belonging to same identities. Each training batch is split into clean $X_{clean}$, globally deformed $X_{global}$ and locally deformed $X_{local}$ subsets. We extract their features and optimize a contrastive loss that brings features of the same identity closer, regardless of the applied transformation:
\[
    \mathcal{L}_{\text{cont}} = 1 - \text{cos\_sim}(F_{\text{global}}, F_{\text{local}}).mean()
\]
where $\text{cos\_sim}(\cdot, \cdot)$ denotes the cosine similarity between two feature vectors. A high similarity (i.e. low contrastive loss) indicates that the network preserves identity information across different deformations.
The total loss used to train the network is a weighted combination of loss on clean samples, the adversarial loss on deformed samples, and the contrastive loss:
\[
\mathcal{L}_{\text{total}} = \mathcal{L}(x,y) +  \mathcal{L}_{\text{trans}}(T_{\psi^*}(x),y) + \lambda_{\text{cont}} \mathcal{L}_{\text{cont}},
\]
where $\mathcal{L}(x,y)$ is the angular loss on clean inputs, $\mathcal{L}(T_{\psi^*}(x),y)$ is the angular loss on deformed inputs, and $\mathcal{L}_{\text{cont}}$ enforces identity consistency across global and local deformations.  The hyperparameter $\lambda_{\text{cont}}$ controls the trade-off between classification accuracy and feature consistency across deformations.

\section{Experiments and Results}
\subsection{Datasets}
\noindent \textbf{Training Dataset.} We trained our models on MS1MV2 \cite{deng2019ArcFace}, MS1MV3 \cite{Deng_2019_ICCV}, VGGFace2 \cite{cao2018vggface2}, WebFace4M and WebFace12M \cite{zhu2021webface260m} datasets. MS1MV2 contains around 5.8M images belonging to around 85K identities, while MS1MV3 contains around 5.1M images and 93K identities. VGGFace2 contains roughly 3.1M images spanning 9K identities. WebFace12M contains 12.7M images and 617K identities. WebFace4M is a subset of WebFace12M with 4.2M images belonging to 205K identities.

\noindent \textbf{Evaluation Dataset.} 
For evaluation we consider three quality categories: low, mixed and high. 
The \textit{low quality} category includes TinyFace dataset \cite{cheng2018low}, the \textit{mixed quality} comprises of IJB-C and IJB-B \cite{whitelam2017iarpa} datasets.  The \textit{high quality} datasets include LFW \cite{huang2008labeled}, CFP-FP \cite{sengupta2016frontal}, AgeDB30 \cite{moschoglou2017agedb}, CALFW \cite{zheng2018cross} and CPLFW \cite{zheng2017cross}.
\begin{table}[t]
  \centering
  \small
  \setlength{\tabcolsep}{6pt}
  \renewcommand{\arraystretch}{1.2}
  \begin{tabular}{lccc}
    \toprule
    \rowcolor{lightgray}
    \textbf{Method} & \textbf{Train Set} &\textbf{Rank 1} & \textbf{Rank 5} \\
    \midrule
    ArcFace \cite{deng2019ArcFace} & MS1MV2  & 63.22 & 67.86 \\
    ArcFace + CFSM \cite{liu2022controllable} & MS1MV2  & 64.69 & 68.80 \\
    ARoFace \cite{saadabadi2024ARoFace} & MS1MV2 & 66.74 & 70.76 \\
    AdaFace \cite{kim2022adaface} & MS1MV2 & 68.21 & 71.54 \\
   \rowcolor{lightred}
    \textbf{DArFace (Ours)} & MS1MV2 & \textbf{68.24} & \textbf{72.26} \\
    \midrule
    ArcFace \cite{deng2019ArcFace} & MS1MV3 & 64.78 & 68.91 \\
    ARoFace \cite{saadabadi2024ARoFace} & MS1MV3 & 68.45 & 72.02 \\
    AdaFace \cite{kim2022adaface} & MS1MV3 & 67.81 & 70.98 \\
    \rowcolor{lightred}
    \textbf{DArFace (Ours)} & MS1MV3 & \textbf{69.39} & \textbf{73.12} \\
    \midrule
    ArcFace \cite{deng2019ArcFace} & VGGFace2 & 63.46 & 69.01 \\
    ARoFace \cite{saadabadi2024ARoFace} & VGGFace2 & 64.69 & 70.25 \\
    \rowcolor{lightred}
    \textbf{DArFace (Ours)} & VGGFace2 & \textbf{65.42} & \textbf{70.70} \\
    \midrule
    ArcFace \cite{deng2019ArcFace} & WebFace4M & 69.68 & 73.28 \\
    ARoFace \cite{saadabadi2024ARoFace} & WebFace4M & 71.62 & 74.67 \\
    AdaFace \cite{kim2022adaface} & WebFace4M & 72.02 & 74.52 \\
    \rowcolor{lightred}
    \textbf{DArFace (Ours)} & WebFace4M & \textbf{72.61} & \textbf{75.59} \\
    \midrule
    ArcFace \cite{deng2019ArcFace} & WebFace12M & 70.25 & 73.20 \\
    ARoFace \cite{saadabadi2024ARoFace} & WebFace12M & 72.71 & 75.61 \\
    AdaFace \cite{kim2022adaface} & WebFace12M & 72.29 & 74.97 \\
    \rowcolor{lightred}
    \textbf{DArFace (Ours)} & WebFace12M & \textbf{73.25} & \textbf{76.01} \\
    \bottomrule
    
  \end{tabular}
  \caption{\textbf{Performance comparison on low-quality (TinyFace) dataset using a ResNet-100 and MS1MV2, MS1MV3 VGGFace2, WebFace4M and WebFace12M training datasets.} Results on TinyFace validate the effectiveness of our proposed global-local deformations under extreme low-quality conditions.}
  \label{tab:lowquality}
\end{table}

\subsection{Implementation Details}
We employ a ResNet-100 backbone with ArcFace loss \cite{deng2019ArcFace}, optimized using SGD (initial LR = 0.2, weight decay = $5e–4$) and a PolynomialLR scheduler for 28 epochs. The framework is compatible with other loss functions.

For deformation-aware augmentation, parameters are sampled from Gaussian distributions:  \( \alpha \sim \mathcal{N}(1, 0.1) \), \( \sigma \sim \mathcal{N}(10, 5) \),  \( s \sim \mathcal{N}(1, 0.01) \),  \( \theta \sim \mathcal{N}(0, 0.01) \), and  \( t \sim \mathcal{N}(0, 0.01) \), with stability bounds \( \alpha \in [0.05, 0.25] \) and \( \sigma \in [20.0, 35.0] \). We set \( \lambda_{\text{cont}} = 0.7 \) and use 75\% clean, 15\% globally transformed, and 10\% locally deformed samples. Following \cite{kim2022adaface}, we also applied crop, low-resolution, and photometric augmentations.

\subsection{High-Quality Images}
To compare with state-of-the-art methods on high quality datasets, we train a ResNet-100 model on the MS1MV2 dataset. As shown in Table. \ref{tab:highquality}, DArFace achieves the highest accuracy across most benchmarks, while performing on par with others. In particular, it achieves 98.62\% on CFP-FP, 96.20\% on CALFW and 93.64\% on CPLFW outperforming ArcFace, MagFace and ARoFace. These demonstrate that DArFace preserves discriminative identity features without compromising performance on clean, high-quality inputs. It is worth noting that performance on these high quality benchmarks is largely saturated in the literature, making further improvements challenging.

\subsection{Mixed-Quality Images}

We evaluate the performance of our method on mixed-quality datasets which include variations in pose, illumination and resolution. To ensure the generality of the improvements, we train DArFace on four different datasets MS1MV2, MS1MV3, WebFace4M, WebFace12M and compare them against the state-of-the-art methods. The performance is reported using  TAR@FAR(True Acceptance Rate at various False Acceptance Rates) and Identification accuracy at Rank 1 and Rank 5.  

As shown in Table. \ref{tab:medquality}, on IJB-B, DArFace trained on MS1MV2 achieves a TAR@FAR of 48.08\% at 1e-6, outperforming MagFace which achieves 42.32\%. While ArcFace+CFSM shows slightly better performance than DArFace, it is important to note that  it is a GAN-based method that requires careful tuning and the generation of synthetic data, adding complexity to the training pipeline. 

When trained on MS1MV3, DArFace consistently outperforms prior methods, including ArcFace+VPL, ArcFace+SC, ArcFace and ARoFace achieving 44.04\% TAR@FAR=1e-6. It also attains strong Rank 1 identification scores of 95.33\% on IJB-B and 96.42\% on IJB-C, highlighting its robustness under mixed-quality conditions.

Training on more challenging WebFace4M and WebFace12M datasets further demonstrate that DArFace maintains comparable or superior performance across both verification and identification metrics, reinforcing its generalization ability across varying data scales and quality levels.

\subsection{Low Quality Images}
To assess the robustness of our method under extreme quality degradations, we evaluate our method on TinyFace dataset, which contains low quality surveillance style face images. We train our model on five different datasets MS1MV2, MS1MV3, VGGFace2, WebFace4M and WebFace12M datasets and report the identification accuracy at Rank 1 and Rank 5. 

As shown in Table. \ref{tab:lowquality}, DArFace achieves the best performance across all the training sets. On MS1MV2 it achieves 68.24\% rank 1 accuracy and 72.26\% rank 5. On MS1MV3 the gains are 69.39\% at rank 1 and 73.12\% at rank 5. Similarly on VGGFace2 Rank-1 accuracy is 65.42\% while Rank-5 is 70.70\%. Compared to ArcFace and ARoFace, DArFace yields consistent gains, demonstrating that the model benefits from explicitly learning robustness to low-quality distortions during training.

When scaled to larger datasets, DArFace  demonstrate consistent gains. On WebFace4M, it reaches 72.61\% at rank 1 and 75.59\% at rank 5, while on the more challenging WebFace12M, it achieves 73.25\% and 76.01\%, surpassing all prior baselines. These results confirm that DArFace effectively generalizes to large-scale data and benefits from its global-local deformation learning, which explicitly enhances robustness to severe low-quality distortions.

\vspace{-2pt}
\begin{table}[h]
  \centering
  \small
  \setlength{\tabcolsep}{5pt}
  \renewcommand{\arraystretch}{1.2}
  \begin{tabular}{lcccc}
    \toprule
    \rowcolor{lightgray}
     \multirow{1}{*}{\textbf{Identification}} & {\textbf{Train Set}} & \textbf{$\mathcal{L}_{clean}$}& 
    \shortstack{\textbf{$\mathcal{L}_{clean}$}\\\textbf{+ $\mathcal{L}_{trans}$}} & 
    \shortstack{\textbf{$\mathcal{L}_{clean}$}\\\textbf{+ $\mathcal{L}_{trans}$}\\\textbf{+ $\mathcal{L}_{cont}$}} \\
    \midrule
    Rank 1  & MS1MV2 & 63.22 & 66.89 & \textbf{67.46} \\
    Rank 5  & MS1MV2 & 67.86 & 71.32 & \textbf{71.45} \\
    Rank 20 & MS1MV2 & 71.35 & 74.30 & \textbf{74.27} \\
    \midrule
    Rank 1  & MS1MV3 & 64.78 & 68.29 & \textbf{68.70} \\
    Rank 5  & MS1MV3 & 68.91 & 71.96 & \textbf{72.59} \\
    Rank 20 & MS1MV3 & 71.99 & 74.49 & \textbf{75.24} \\
    \bottomrule
  \end{tabular}
  \vspace{2pt}
  \caption{\textbf{Evaluating the impact of contrastive loss ($\mathcal{L}_{cont}$) on the TinyFace dataset using a ResNet-100 backbone.} Incorporating $\mathcal{L}_{trans}$ improves accuracy over the clean baseline, while adding $\mathcal{L}_{cont}$ yields further gains, especially at higher ranks demonstrating the benefit of learning consistent representations under global and local deformations.}
  \label{tab:cont_loss}
\end{table}
\begin{table*}
  \centering
\begin{tabular}{c c c c c c c}
\toprule
\rowcolor{lightgray}
 Method &  Resolution & LFW & CFP-FP & AgeDB30 & CALFW & CPLFW \\
\midrule
\multirow{3}{*}{ArcFace \cite{deng2019ArcFace}} & $8 \times 8$ &71.96 &\textbf{59.65} &54.18 &59.30 &58.88\\ & $16 \times 16$ & 91.68 &73.05 & 68.09&74.36 & 74.81 \\ & Original &99.76 & 98.54 & \textbf{98.28} & 96.10 & 93.16\\
\midrule
\multirow{3}{*}{ARoFace \cite{saadabadi2024ARoFace}} & $8 \times 8$ & 71.50 & 59.35 &55.30 & 58.25 &59.51\\ & $16 \times 16$ & 91.40 & 73.25 & 67.28 & 74.73 & 74.03\\ & Original & 99.78 & 98.51 & 97.80 & 95.80 & 93.43\\
\midrule
\multirow{3}{*}{\textbf{DArFace (Ours)}}& $8 \times 8$ & \textbf{73.50} & 59.58 &\textbf{55.61} &\textbf{60.45} &\textbf{60.28}\\ 
& $16 \times 16$ & \textbf{95.10} & \textbf{81.04} &\textbf{73.31} &\textbf{81.09} &\textbf{80.63}\\
& Original & \textbf{99.80} & \textbf{98.62} & 98.06 & \textbf{96.20} & \textbf{93.64} \\

\bottomrule
\end{tabular}
\vspace{2pt}
\caption{ \textbf{Performance comparison on down-sampled inputs across multiple face verification benchmarks.} DArFace consistently achieves the best performance, particularly at 
$16 \times16$ and $8 \times 8$ resolutions, highlighting its robustness to severe quality degradations. }
\label{tab:downsampled}
\end{table*}
\begin{table}
  \centering
  \small
  \setlength{\tabcolsep}{6pt}
  \renewcommand{\arraystretch}{1.2}
  \begin{tabular}{lccc}
    \toprule
    \rowcolor{lightgray}
    \textbf{Identification} & \textbf{Train Set} & 
    \shortstack{\textbf{DArFace}\\\textbf{w/o Aug}} & 
    \shortstack{\textbf{DArFace}\\\textbf{with Aug}} \\
    \midrule
    Rank-1  & MS1MV2 & 67.46 & \textbf{68.24} \\
    Rank-5  & MS1MV2 & 71.45 & \textbf{72.26} \\
    Rank-20 & MS1MV2 & 74.27 & \textbf{74.97} \\
    \midrule
    Rank-1  & MS1MV3 & 68.70 & \textbf{69.39} \\
    Rank-5  & MS1MV3 & 72.59 & \textbf{73.12} \\
    Rank-20 & MS1MV3 & 75.24 & \textbf{75.45} \\

    \bottomrule
  \end{tabular}
  \vspace{2pt}
  \caption{\textbf{Evaluating the effect of augmentation strategies on low-quality TinyFace dataset using a ResNet-100 backbone.} Adding crop, low resolution and photometric augmentations during training leads to consistent improvements across all identification ranks.}
  \label{tab:aug}
\end{table}
\vspace{-2pt}
\subsection{Ablations}
\noindent \textbf{Effect of Contrastive Loss.}
We analyze the contribution of each loss component used in our training framework. In Table. \ref{tab:cont_loss}, $\mathcal{L}_{clean}$ denotes training with only clean inputs using an angular loss, incorporating $\mathcal{L}_{trans}$ introduces both global and local transformations into the training process, encouraging the model to learn transformation-robust representations. The contrastive loss $\mathcal{L}_{cont}$ further aligns different deformed views to promote consistency. Models are trained on MS1MV2 and MS1MV3 datasets and evaluated on the very low-quality TinyFace dataset. We observe that adding deformation-based training by $\mathcal{L}_{trans}$ enhances the performance by around 3\% across datasets. Integrating contrastive supervision through $\mathcal{L}_{cont}$ yields further improvements of around 1\%, with largest gains observed in the higher ranks. This demonstrates the effectiveness of aligning feature representations across different deformed views, allowing the model to learn more consistent and deformation-invariant identity features.

\noindent \textbf{Effect of Augmentation Strategies.} Following \cite{kim2022adaface}, we apply random cropping, low-resolution simulation, and photometric transformations during training, each with a probability of 0.2. Models are trained on MS1MV2 and MS1MV3, and evaluated on the low-quality TinyFace dataset. As shown in Table~\ref{tab:aug}, these augmentations consistently improve performance across all identification ranks, demonstrating their effectiveness in enhancing model robustness to image degradations.

\noindent \textbf{Effect of Down Sampling.} To further evaluate the robustness of our method, we conduct a resolution-based down sampling ablation. We evaluate our model on face verification benchmarks by down sampling them to $8\times8$ and $16\times16$ pixels. Models are trained on MS1MV2 dataset. 

As in Table. \ref{tab:downsampled}, DArFace outperforms previous baselines across all resolutions and datasets. For LFW and CPLFW datasets at the resolution of $8\times8$ the gain is around 2\%. For $16\times16$ the recovery by DArFace is even more significant with 81.04\% with DArFace and around 73\% with ARoFace and ArcFace for CFP-FP dataset. Similarly, For AgeDB30 DArFace shows 73.31\% while ArcFace and ARoFace are 68.09\% and 67.28\% respectively. These results demonstrate the robustness of DArFace under severe resolution degradations, highlighting its ability to recover discriminative facial features even from $8\times8$ or $16\times16$ inputs.
\vspace{-4pt}
\section{Conclusions}
This paper introduces \textbf{DArFace}, a novel deformation-aware training framework for low-quality face recognition. By incorporating both global (e.g., rotation, scaling) and local (e.g., elastic distortions) deformations during training, and leveraging a contrastive identity loss to enforce feature consistency between these variations, DArFace enhances the model’s robustness to the real-world quality degradations. Extensive experiments and ablation studies across multiple datasets demonstrate the effectiveness of our approach, showing consistent improvements over existing methods. DArFace not only preserves discriminative identity features under clean conditions but also significantly boosts performance on challenging, low-quality benchmarks.
\vspace{-4pt}
{\small
\bibliographystyle{ieee}
\bibliography{egbib}

\begin{thebibliography}{10}\itemsep=-1pt

\bibitem{cao2018vggface2}
Q.~Cao, L.~Shen, W.~Xie, O.~M. Parkhi, and A.~Zisserman.
\newblock Vggface2: A dataset for recognising faces across pose and age.
\newblock In {\em 2018 13th IEEE international conference on automatic face \& gesture recognition (FG 2018)}, pages 67--74. IEEE, 2018.

\bibitem{Chang_2020_CVPR}
J.~Chang, Z.~Lan, C.~Cheng, and Y.~Wei.
\newblock Data uncertainty learning in face recognition.
\newblock In {\em Proceedings of the IEEE/CVF Conference on Computer Vision and Pattern Recognition (CVPR)}, June 2020.

\bibitem{cheng2018low}
Z.~Cheng, X.~Zhu, and S.~Gong.
\newblock Low-resolution face recognition.
\newblock In {\em Asian conference on computer vision}, pages 605--621. Springer, 2018.

\bibitem{deb2020advfaces}
D.~Deb, J.~Zhang, and A.~K. Jain.
\newblock Advfaces: Adversarial face synthesis.
\newblock In {\em 2020 IEEE International Joint Conference on Biometrics (IJCB)}, pages 1--10. IEEE, 2020.

\bibitem{deng2020sub}
J.~Deng, J.~Guo, T.~Liu, M.~Gong, and S.~Zafeiriou.
\newblock Sub-center arcface: Boosting face recognition by large-scale noisy web faces.
\newblock In {\em Computer Vision--ECCV 2020: 16th European Conference, Glasgow, UK, August 23--28, 2020, Proceedings, Part XI 16}, pages 741--757. Springer, 2020.

\bibitem{deng2019ArcFace}
J.~Deng, J.~Guo, N.~Xue, and S.~Zafeiriou.
\newblock Arcface: Additive angular margin loss for deep face recognition.
\newblock In {\em Proceedings of the IEEE/CVF conference on computer vision and pattern recognition}, pages 4690--4699, 2019.

\bibitem{deng2021variational}
J.~Deng, J.~Guo, J.~Yang, A.~Lattas, and S.~Zafeiriou.
\newblock Variational prototype learning for deep face recognition.
\newblock In {\em Proceedings of the IEEE/CVF Conference on Computer Vision and Pattern Recognition}, pages 11906--11915, 2021.

\bibitem{Deng_2019_ICCV}
J.~Deng, J.~Guo, D.~Zhang, Y.~Deng, X.~Lu, and S.~Shi.
\newblock Lightweight face recognition challenge.
\newblock In {\em Proceedings of the IEEE/CVF International Conference on Computer Vision (ICCV) Workshops}, Oct 2019.

\bibitem{ge2020efficient}
S.~Ge, S.~Zhao, C.~Li, Y.~Zhang, and J.~Li.
\newblock Efficient low-resolution face recognition via bridge distillation.
\newblock {\em IEEE Transactions on Image Processing}, 29:6898--6908, 2020.

\bibitem{goodfellow2014explaining}
I.~J. Goodfellow, J.~Shlens, and C.~Szegedy.
\newblock Explaining and harnessing adversarial examples.
\newblock {\em arXiv preprint arXiv:1412.6572}, 2014.

\bibitem{guo2016ms}
Y.~Guo, L.~Zhang, Y.~Hu, X.~He, and J.~Gao.
\newblock Ms-celeb-1m: A dataset and benchmark for large-scale face recognition.
\newblock In {\em Computer Vision--ECCV 2016: 14th European Conference, Amsterdam, The Netherlands, October 11-14, 2016, Proceedings, Part III 14}, pages 87--102. Springer, 2016.

\bibitem{he2016deep}
K.~He, X.~Zhang, S.~Ren, and J.~Sun.
\newblock Deep residual learning for image recognition.
\newblock In {\em Proceedings of the IEEE conference on computer vision and pattern recognition}, pages 770--778, 2016.

\bibitem{hinton2011transforming}
G.~E. Hinton, A.~Krizhevsky, and S.~D. Wang.
\newblock Transforming auto-encoders.
\newblock In {\em Artificial Neural Networks and Machine Learning--ICANN 2011: 21st International Conference on Artificial Neural Networks, Espoo, Finland, June 14-17, 2011, Proceedings, Part I 21}, pages 44--51. Springer, 2011.

\bibitem{hoffer2015deep}
E.~Hoffer and N.~Ailon.
\newblock Deep metric learning using triplet network.
\newblock In {\em Similarity-based pattern recognition: third international workshop, SIMBAD 2015, Copenhagen, Denmark, October 12-14, 2015. Proceedings 3}, pages 84--92. Springer, 2015.

\bibitem{huang2017densely}
G.~Huang, Z.~Liu, L.~Van Der~Maaten, and K.~Q. Weinberger.
\newblock Densely connected convolutional networks.
\newblock In {\em Proceedings of the IEEE conference on computer vision and pattern recognition}, pages 4700--4708, 2017.

\bibitem{huang2008labeled}
G.~B. Huang, M.~Mattar, T.~Berg, and E.~Learned-Miller.
\newblock Labeled faces in the wild: A database forstudying face recognition in unconstrained environments.
\newblock In {\em Workshop on faces in'Real-Life'Images: detection, alignment, and recognition}, 2008.

\bibitem{Huang_2017_ICCV}
R.~Huang, S.~Zhang, T.~Li, and R.~He.
\newblock Beyond face rotation: Global and local perception gan for photorealistic and identity preserving frontal view synthesis.
\newblock In {\em Proceedings of the IEEE International Conference on Computer Vision (ICCV)}, Oct 2017.

\bibitem{huang2024enhanced}
Y.~Huang, Y.~Wang, L.~Yang, and L.~Wang.
\newblock Enhanced face recognition using intra-class incoherence constraint.
\newblock In {\em The Twelfth International Conference on Learning Representations}, 2024.

\bibitem{kalka2018ijb}
N.~D. Kalka, B.~Maze, J.~A. Duncan, K.~O’Connor, S.~Elliott, K.~Hebert, J.~Bryan, and A.~K. Jain.
\newblock Ijb--s: Iarpa janus surveillance video benchmark.
\newblock In {\em 2018 IEEE 9th international conference on biometrics theory, applications and systems (BTAS)}, pages 1--9. IEEE, 2018.

\bibitem{kemelmacher2016megaface}
I.~Kemelmacher-Shlizerman, S.~M. Seitz, D.~Miller, and E.~Brossard.
\newblock The megaface benchmark: 1 million faces for recognition at scale.
\newblock In {\em Proceedings of the IEEE conference on computer vision and pattern recognition}, pages 4873--4882, 2016.

\bibitem{kim2022adaface}
M.~Kim, A.~K. Jain, and X.~Liu.
\newblock Adaface: Quality adaptive margin for face recognition.
\newblock In {\em Proceedings of the IEEE/CVF conference on computer vision and pattern recognition}, pages 18750--18759, 2022.

\bibitem{komkov2021advhat}
S.~Komkov and A.~Petiushko.
\newblock Advhat: Real-world adversarial attack on arcface face id system.
\newblock In {\em 2020 25th international conference on pattern recognition (ICPR)}, pages 819--826. IEEE, 2021.

\bibitem{krizhevsky2012imagenet}
A.~Krizhevsky, I.~Sutskever, and G.~E. Hinton.
\newblock Imagenet classification with deep convolutional neural networks.
\newblock {\em Advances in neural information processing systems}, 25, 2012.

\bibitem{kurakin2018adversarial}
A.~Kurakin, I.~J. Goodfellow, and S.~Bengio.
\newblock Adversarial examples in the physical world.
\newblock In {\em Artificial intelligence safety and security}, pages 99--112. Chapman and Hall/CRC, 2018.

\bibitem{Li_2021_CVPR}
S.~Li, J.~Xu, X.~Xu, P.~Shen, S.~Li, and B.~Hooi.
\newblock Spherical confidence learning for face recognition.
\newblock In {\em Proceedings of the IEEE/CVF Conference on Computer Vision and Pattern Recognition (CVPR)}, pages 15629--15637, June 2021.

\bibitem{Li_2020_CVPR}
X.~Li, W.~Li, D.~Ren, H.~Zhang, M.~Wang, and W.~Zuo.
\newblock Enhanced blind face restoration with multi-exemplar images and adaptive spatial feature fusion.
\newblock In {\em Proceedings of the IEEE/CVF Conference on Computer Vision and Pattern Recognition (CVPR)}, June 2020.

\bibitem{liu2022controllable}
F.~Liu, M.~Kim, A.~Jain, and X.~Liu.
\newblock Controllable and guided face synthesis for unconstrained face recognition.
\newblock In {\em European Conference on Computer Vision}, pages 701--719. Springer, 2022.

\bibitem{liu2017sphereface}
W.~Liu, Y.~Wen, Z.~Yu, M.~Li, B.~Raj, and L.~Song.
\newblock Sphereface: Deep hypersphere embedding for face recognition.
\newblock In {\em Proceedings of the IEEE conference on computer vision and pattern recognition}, pages 212--220, 2017.

\bibitem{madry2017towards}
A.~Madry, A.~Makelov, L.~Schmidt, D.~Tsipras, and A.~Vladu.
\newblock Towards deep learning models resistant to adversarial attacks.
\newblock {\em arXiv preprint arXiv:1706.06083}, 2017.

\bibitem{maze2018iarpa}
B.~Maze, J.~Adams, J.~A. Duncan, N.~Kalka, T.~Miller, C.~Otto, A.~K. Jain, W.~T. Niggel, J.~Anderson, J.~Cheney, et~al.
\newblock Iarpa janus benchmark-c: Face dataset and protocol.
\newblock In {\em 2018 international conference on biometrics (ICB)}, pages 158--165. IEEE, 2018.

\bibitem{meng2021magface}
Q.~Meng, S.~Zhao, Z.~Huang, and F.~Zhou.
\newblock Magface: A universal representation for face recognition and quality assessment.
\newblock In {\em Proceedings of the IEEE/CVF conference on computer vision and pattern recognition}, pages 14225--14234, 2021.

\bibitem{moosavi2016deepfool}
S.-M. Moosavi-Dezfooli, A.~Fawzi, and P.~Frossard.
\newblock Deepfool: a simple and accurate method to fool deep neural networks.
\newblock In {\em Proceedings of the IEEE conference on computer vision and pattern recognition}, pages 2574--2582, 2016.

\bibitem{moschoglou2017agedb}
S.~Moschoglou, A.~Papaioannou, C.~Sagonas, J.~Deng, I.~Kotsia, and S.~Zafeiriou.
\newblock Agedb: the first manually collected, in-the-wild age database.
\newblock In {\em proceedings of the IEEE conference on computer vision and pattern recognition workshops}, pages 51--59, 2017.

\bibitem{5204299}
N.~Ozay, Y.~Tong, F.~W. Wheeler, and X.~Liu.
\newblock Improving face recognition with a quality-based probabilistic framework.
\newblock In {\em 2009 IEEE Computer Society Conference on Computer Vision and Pattern Recognition Workshops}, pages 134--141, 2009.

\bibitem{papernot2016limitations}
N.~Papernot, P.~McDaniel, S.~Jha, M.~Fredrikson, Z.~B. Celik, and A.~Swami.
\newblock The limitations of deep learning in adversarial settings.
\newblock In {\em 2016 IEEE European symposium on security and privacy (EuroS\&P)}, pages 372--387. IEEE, 2016.

\bibitem{saadabadi2024ARoFace}
M.~S.~E. Saadabadi, S.~R. Malakshan, A.~Dabouei, and N.~M. Nasrabadi.
\newblock Aroface: Alignment robustness to improve low-quality face recognition.
\newblock In {\em European Conference on Computer Vision}, pages 308--327. Springer, 2024.

\bibitem{schroff2015facenet}
F.~Schroff, D.~Kalenichenko, and J.~Philbin.
\newblock Facenet: A unified embedding for face recognition and clustering.
\newblock In {\em Proceedings of the IEEE conference on computer vision and pattern recognition}, pages 815--823, 2015.

\bibitem{sengupta2016frontal}
S.~Sengupta, J.-C. Chen, C.~Castillo, V.~M. Patel, R.~Chellappa, and D.~W. Jacobs.
\newblock Frontal to profile face verification in the wild.
\newblock In {\em 2016 IEEE winter conference on applications of computer vision (WACV)}, pages 1--9. IEEE, 2016.

\bibitem{sharif2016accessorize}
M.~Sharif, S.~Bhagavatula, L.~Bauer, and M.~K. Reiter.
\newblock Accessorize to a crime: Real and stealthy attacks on state-of-the-art face recognition.
\newblock In {\em Proceedings of the 2016 acm sigsac conference on computer and communications security}, pages 1528--1540, 2016.

\bibitem{sharif2019general}
M.~Sharif, S.~Bhagavatula, L.~Bauer, and M.~K. Reiter.
\newblock A general framework for adversarial examples with objectives.
\newblock {\em ACM Transactions on Privacy and Security (TOPS)}, 22(3):1--30, 2019.

\bibitem{shi2021boosting}
Y.~Shi and A.~K. Jain.
\newblock Boosting unconstrained face recognition with auxiliary unlabeled data.
\newblock In {\em Proceedings of the IEEE/CVF Conference on Computer Vision and Pattern Recognition}, pages 2795--2804, 2021.

\bibitem{shin2022teaching}
S.~Shin, J.~Lee, J.~Lee, Y.~Yu, and K.~Lee.
\newblock Teaching where to look: Attention similarity knowledge distillation for low resolution face recognition.
\newblock In {\em European Conference on Computer Vision}, pages 631--647. Springer, 2022.

\bibitem{1227801}
P.~Simard, D.~Steinkraus, and J.~Platt.
\newblock Best practices for convolutional neural networks applied to visual document analysis.
\newblock In {\em Seventh International Conference on Document Analysis and Recognition, 2003. Proceedings.}, pages 958--963, 2003.

\bibitem{singh2019dual}
M.~Singh, S.~Nagpal, R.~Singh, and M.~Vatsa.
\newblock Dual directed capsule network for very low resolution image recognition.
\newblock In {\em Proceedings of the IEEE/CVF international conference on computer vision}, pages 340--349, 2019.

\bibitem{szegedy2013intriguing}
C.~Szegedy, W.~Zaremba, I.~Sutskever, J.~Bruna, D.~Erhan, I.~Goodfellow, and R.~Fergus.
\newblock Intriguing properties of neural networks.
\newblock {\em arXiv preprint arXiv:1312.6199}, 2013.

\bibitem{wang2018cosface}
H.~Wang, Y.~Wang, Z.~Zhou, X.~Ji, D.~Gong, J.~Zhou, Z.~Li, and W.~Liu.
\newblock Cosface: Large margin cosine loss for deep face recognition.
\newblock In {\em Proceedings of the IEEE conference on computer vision and pattern recognition}, pages 5265--5274, 2018.

\bibitem{wang2019improved}
M.~Wang, R.~Liu, N.~Hajime, A.~Narishige, H.~Uchida, and T.~Matsunami.
\newblock Improved knowledge distillation for training fast low resolution face recognition model.
\newblock In {\em Proceedings of the IEEE/CVF International Conference on Computer Vision Workshops}, pages 0--0, 2019.

\bibitem{Wang_2022_CVPR}
Z.~Wang, J.~Zhang, R.~Chen, W.~Wang, and P.~Luo.
\newblock Restoreformer: High-quality blind face restoration from undegraded key-value pairs.
\newblock In {\em Proceedings of the IEEE/CVF Conference on Computer Vision and Pattern Recognition (CVPR)}, pages 17512--17521, June 2022.

\bibitem{whitelam2017iarpa}
C.~Whitelam, E.~Taborsky, A.~Blanton, B.~Maze, J.~Adams, T.~Miller, N.~Kalka, A.~K. Jain, J.~A. Duncan, K.~Allen, et~al.
\newblock Iarpa janus benchmark-b face dataset.
\newblock In {\em proceedings of the IEEE conference on computer vision and pattern recognition workshops}, pages 90--98, 2017.

\bibitem{xiao2018spatially}
C.~Xiao, J.-Y. Zhu, B.~Li, W.~He, M.~Liu, and D.~Song.
\newblock Spatially transformed adversarial examples.
\newblock {\em arXiv preprint arXiv:1801.02612}, 2018.

\bibitem{10.1007/978-3-319-10593-2_25}
C.-Y. Yang, C.~Ma, and M.-H. Yang.
\newblock Single-image super-resolution: A benchmark.
\newblock In {\em Computer Vision -- ECCV 2014}, pages 372--386, Cham, 2014. Springer International Publishing.

\bibitem{yang2021attacks}
L.~Yang, Q.~Song, and Y.~Wu.
\newblock Attacks on state-of-the-art face recognition using attentional adversarial attack generative network.
\newblock {\em Multimedia tools and applications}, 80:855--875, 2021.

\bibitem{yang2021gan}
T.~Yang, P.~Ren, X.~Xie, and L.~Zhang.
\newblock Gan prior embedded network for blind face restoration in the wild.
\newblock In {\em Proceedings of the IEEE/CVF conference on computer vision and pattern recognition}, pages 672--681, 2021.

\bibitem{8723565}
W.~Yang, X.~Zhang, Y.~Tian, W.~Wang, J.-H. Xue, and Q.~Liao.
\newblock Deep learning for single image super-resolution: A brief review.
\newblock {\em IEEE Transactions on Multimedia}, 21(12):3106--3121, 2019.

\bibitem{zangeneh2020low}
E.~Zangeneh, M.~Rahmati, and Y.~Mohsenzadeh.
\newblock Low resolution face recognition using a two-branch deep convolutional neural network architecture.
\newblock {\em Expert Systems with Applications}, 139:112854, 2020.

\bibitem{ZANGENEH2020112854}
E.~Zangeneh, M.~Rahmati, and Y.~Mohsenzadeh.
\newblock Low resolution face recognition using a two-branch deep convolutional neural network architecture.
\newblock {\em Expert Systems with Applications}, 139:112854, 2020.

\bibitem{zheng2018cross}
T.~Zheng and W.~Deng.
\newblock Cross-pose lfw: A database for studying cross-pose face recognition in unconstrained environments.
\newblock {\em Beijing University of Posts and Telecommunications, Tech. Rep}, 5(7):5, 2018.

\bibitem{zheng2017cross}
T.~Zheng, W.~Deng, and J.~Hu.
\newblock Cross-age lfw: A database for studying cross-age face recognition in unconstrained environments.
\newblock {\em arXiv preprint arXiv:1708.08197}, 2017.

\bibitem{zhu2021webface260m}
Z.~Zhu, G.~Huang, J.~Deng, Y.~Ye, J.~Huang, X.~Chen, J.~Zhu, T.~Yang, J.~Lu, D.~Du, et~al.
\newblock Webface260m: A benchmark unveiling the power of million-scale deep face recognition.
\newblock In {\em Proceedings of the IEEE/CVF Conference on Computer Vision and Pattern Recognition}, pages 10492--10502, 2021.

\bibitem{zou2011very}
W.~W. Zou and P.~C. Yuen.
\newblock Very low resolution face recognition problem.
\newblock {\em IEEE Transactions on image processing}, 21(1):327--340, 2011.

\end{thebibliography}
}

\end{document}